\documentclass[twocolumn,10pt]{asme2e}

\usepackage{graphicx}
\usepackage{subcaption}
\usepackage{float}
\usepackage{amsmath}
\usepackage{color,soul}
\usepackage{fancyhdr}
\usepackage{lipsum}

\fancypagestyle{alim}{\renewcommand{\headrulewidth}{0pt}\fancyfoot{  This material is based upon work supported by DARPA under Contract No. HR0011-18-3-0010. Any opinions, findings and conclusions or recommendations expressed in this material are those of the author(s) and do not necessarily reflect the views of the U.S. Government. Distribution Statement "A" (Approved for Public Release, Distribution Unlimited).}}

\usepackage[ruled, linesnumbered]{algorithm2e}

\confshortname{IDETC/DAC 2020}
\conffullname{the ASME 2020 International Design Engineering Technical Conferences \&\\
              Computers and Information in Engineering Conference}

\confdate{August}
\confyear{2020}
\confcity{St. Louis}
\confcountry{USA}

\papernum{IDETC2020-19380}

\title{Attention Routing: track-assignment detailed routing using attention-based Reinforcement Learning}

\author{HAIGUANG LIAO$^1$ \hspace{12pt}
       {\tensfb QINGYI DONG$^1$} \hspace{12pt}
       {\tensfb XULIANG DONG$^1$} \hspace{12pt}
       {\tensfb WENTAI ZHANG$^1$} \hspace{12pt}
       \\  
       {\tensfb WANGYANG ZHANG$^2$} \hspace{12pt}
       {\tensfb WEIYI QI$^2$} \hspace{12pt}
       {\tensfb ELIAS FALLON$^2$} \hspace{12pt}
       {\tensfb LEVENT BURAK KARA$^1$}\thanks{Address all correspondence to this author.}
       \affiliation{
	1. Carnegie Mellon University\\
	Pittsburgh, PA 15213\\
	2. Cadence Design Systems\\
	San Jose, CA 95134
    }
}

\fancypagestyle{alim}{\fancyhf{}\renewcommand{\headrulewidth}{0pt}\fancyfoot[C]{\fontsize{8}{8} \selectfont  This material is based upon work supported by DARPA under Contract No. HR0011-18-3-0010. Any opinions, findings and conclusions or recommendations expressed in this material are those of the author(s) and do not necessarily reflect the views of the U.S. Government. Distribution Statement "A" (Approved for Public Release, Distribution Unlimited).The views, opinions and/or findings expressed are those of the author and should not be interpreted as representing the official views or policies of the Department of Defense or the U.S. Government.}}

\begin{document}
\fancyhead{}
\renewcommand{\headrulewidth}{0pt}

\maketitle    
\thispagestyle{alim}
\begin{abstract}
{\it In the physical design of integrated circuits, global and detailed routing are critical stages involving the determination of the interconnected paths of each net on a circuit while satisfying the design constraints. Existing actual routers as well as routability predictors either have to resort to expensive approaches that lead to high computational times, or use heuristics that do not generalize well. Even though new, learning-based routing methods have been proposed to address this need, requirements on labelled data and difficulties in addressing complex design rule constraints have limited their adoption in advanced technology node physical design problems. In this work, we propose a new router --- attention router, which is the first attempt to solve the track-assignment detailed routing problem \textcolor{black}{ by applying} reinforcement learning. Complex design rule constraints are encoded into the routing algorithm and an attention-model-based REINFORCE algorithm is applied to solve the most critical step in routing: sequencing device pairs to be routed. The attention router and its baseline genetic router are applied to solve different commercial advanced technologies analog circuits problem sets. The attention router demonstrates generalization ability to  unseen problems and is also able to achieve more than $100 \times$  acceleration over the genetic router without \textcolor{black}{severely} compromising the routing solution quality. \textcolor{black}{Increasing the number of training problems greatly improves the performance of attention router.} We also discover a similarity between the attention router and the baseline genetic router in terms of positive correlations in cost and  routing patterns, which demonstrate the attention router's ability to be utilized not only as a detailed router but also as a predictor for routability and congestion.}

\end{abstract}


\section{INTRODUCTION}
Integrated circuits (IC) are becoming increasingly more sophisticated keeping pace with Moore's Law \cite{schaller1997moore}. To solve the increasingly more complex IC system design problems, new advanced electronic design automation (EDA) tools are needed to help engineers  especially in the domain of advanced technology node ($<$ 16 nm) IC designs. In the physical design flow of IC, a critical step is \textit{routing}, where  paths for connecting separate groups of devices are generated based on the locations of devices determined in the previous step of placement. To make the problem tractable, the routing problem is  addressed in two stages: \textit{global routing} and \textit{detailed routing} \cite{sherwani2012algorithms}. While global routing aims to coarsely assign space resources used for routing, detailed routing generates the exact routes that connnect electric components. The placement and routing, while applied sequentially,  are interdependent: a good placement  makes routing  simpler, and quantitative routing measures can in turn  be used to assess the quality of a placement solution.

To achieve successful and high quality IC physical designs, prior works have emphasized quality of routing \cite{hu2001survey,mo2001force,cong1992provably} vs. speed of attaining a solution \cite{soukup1978fast,chang2008nthu}, and have developed routability prediction algorithms \cite{li1999routability,pui2017clock}.  However, existing routing algorithms are primarily based on heuristic based methods which impose stringent constraints and therefore do not generalize well to unseen problems. Although, there have been several learning-based methods   to improve the performance of routing algorithms \cite{qi2014accurate,tabrizi2018machine,liao2020deep}, these approaches either only work as routability predictors or are hampered by  limited generalization ability and the inability to account for complex design rule constraints, which are becoming increasingly sophisticated in advanced technology nodes IC design. As such, fast routing algorithms with strong generalization ability are urgently needed. 
In this work, we present \textit{attention routing}, which is an \textcolor{black}{application of} attention-model based reinforcement learning (RL) model, to solve the track-assignment detailed routing problems on  advanced node technologies problem sets. The routing algorithm  is designed to encode the design rules into the track-assignment steps. The RL algorithm addresses one of the most critical steps in  routing, which is determining the best order sequence of the set of device pairs to be routed, such that the overall solution quality is maximized. To the best of our knowledge, this work is the first attempt to solve the detailed routing problem using RL. The RL model is a policy gradient method based on attention model \cite{kool2018attention}. We describe our attention router and also a genetic router, which is based on genetic algorithms (GA). Both methods are tested on commercial advanced technology nodes IC problem sets, performance is compared and analyzed.

\section{BACKGROUND and RELATED WORK}
\subsection{Width Spacing Pattern}
In advanced technologies node, the manufacturing and design rule constraints (\textit{e.g.} those due to multi-patterning) have significantly increased the complexity of the physical design task. As a result, it is becoming increasingly more challenging for layout designers to parse and memorize all the design rules. A further layer of abstraction is introduced to address this issue, namely the Width Spacing Pattern (WSP).

WSPs define a set of track patterns that consist of different width and spacing configurations for metal wires. By restricting the routes on the WSP rows and tracks, many design rules associated with full custom designs, including those concerning the spacing, minimum widths, and coloring rules can be avoided. In this work, as we attempt to solve the detailed routing problem for analog circuits in advanced technologies (FinFET), the routing strategies we present  follow  the design rule specifications through the adoption of WSP abstraction.

\subsection{Track-Assignment Routing}
As mentioned above, routing is typically divided into two stages: \textit{global routing} and \textit{detailed routing} \cite{sherwani2012algorithms}. In  global routing, the routing resources are allocated into sub-regions (in the WSP setting, rows) and the detailed router will then implement the planned routes, satisfying various constraints (\textit{e.g.} open, short, design rule checkers (DRC). However, when many routes are sequentially implemented, the two-step solution could result in undesirable detours for global routes that are planned to be straight \cite{hetzel1998sequential}. Such issues can be addressed with a time-consuming rip-up and re-route strategy, which involves  heuristics that depend on the routing style and technology requirements \cite{chen2009global}. Another approach is to insert a track assignment step between the global and detailed routing stages that aims to solve the routing problem in a more hierarchical manner \cite{sriram1992detailed, zhou1999global, batterywala2002track, wu2004layer, liu2010global}.

\begin{figure}[thpb]
\centering
\includegraphics[width=0.46\textwidth]{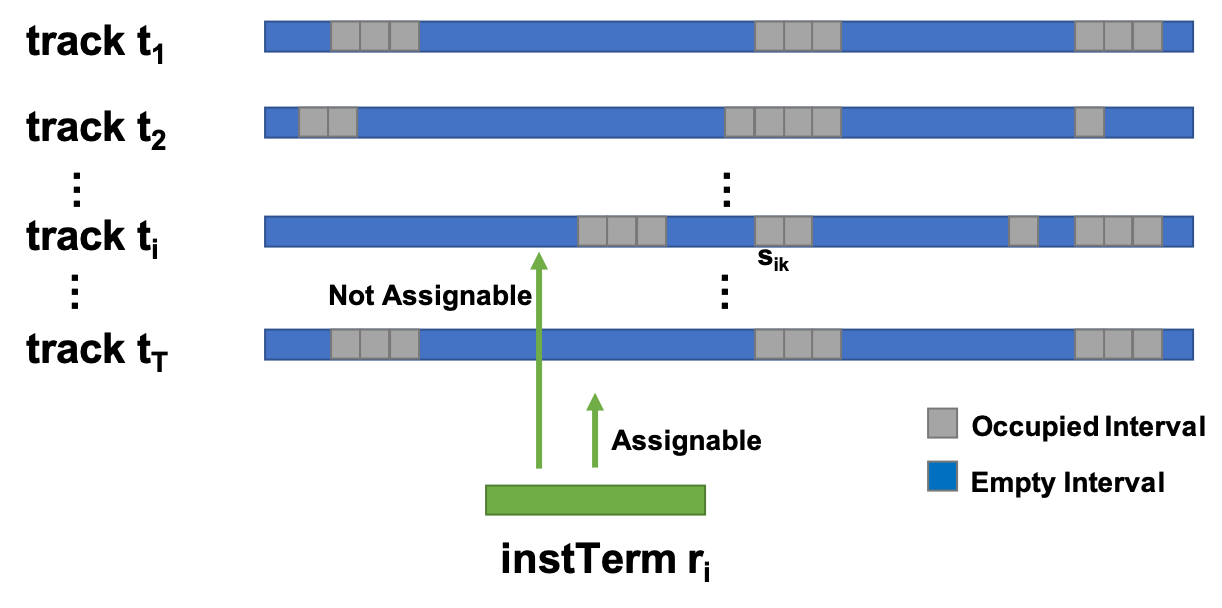}
\caption{Schematic showing track assignment constraints.}
\label{trackinstTerm}
\end{figure}

The goal of track assignment is to place the long routes onto tracks defined by the WSP, with the constraints imposed by technology, routing resources, as well as conflicting routes being simultaneously considered. It also facilitates addressing layout dependent issues such as crosstalk \cite{wu2004layer}. After the long routes are embedded on the tracks, the detailed router's job is to connect those components (instance terminals, instTerms) belonging to the same net, thereby significantly reducing the routing search space. 

Let us denote $t_i$ as the $i$-th track, ${T}$ the set of tracks defined by the WSP, and $s_{ik}$ the $k$-th occupied interval on track $t_i$, as shown in Fig.~\ref{trackinstTerm}. Then, the utilized track resources on $t_i$ is: $u_{t_i} = \bigcup_ks_{ik}$, and instTerm $r_i$ is \textit{assignable} to track $t_j$ \textit{iff} the respective track interval is not occupied ($r_i \cap u_{t_i} = \emptyset$). Similarly, the instTerms to be assigned are defined as, $r_i \in {I}$, where $r_i$ is the $i$-th instTerms and ${I}$ the instTerm set extracted from the global routing result. The task of assigning instTerm $r_i$ onto track $t_j$, denoted as $m_{ij}$, is associated with an assignment cost $C_{ij}$, which reflects the cost including track occupation, perpendicular connection, via insertion \cite{batterywala2002track}. The track-assignment step is therefore deciding a mapping ${M}$ which assigns all the instTerms onto the available tracks without any conflict, while minimizing the assignment cost:

\begin{equation}
\label{trackassignment}
    M^* = \min_{M}\left\{\sum_{m_{ij} \in M} C_{ij}\right\}
\end{equation}
\textit{s.t.} $\forall m_{ij} \in M^*$, $r_i \cap u_{t_i} = \emptyset$

Note that this is  a modified weighted bipartite matching problem, which is known to be NP-complete. As in \cite{batterywala2002track}, we solve it using a heuristic based algorithm, and the details are discussed in Section \ref{track_assignment_section}.

\begin{figure}[h]
\includegraphics[width=0.45\textwidth]{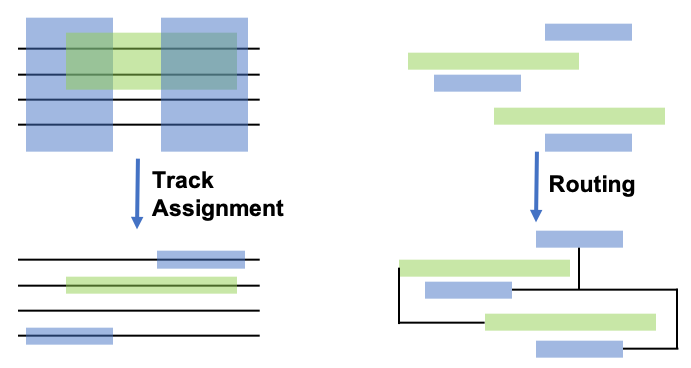}
\caption{Schematic showing track assignment and routing.}
\label{track-assignment} 
\end{figure}

More specifically, in our case (as shown in Figure~\ref{track-assignment}), the routing task consists of two sub-tasks, \textit{i.e.} routing the instTerms on the appropriate tracks and connecting the instTerms. Although an instTerm could consist of many pins with different x-coordinates, an instTerm must be routed on a \textit{single} track, making it suitable to use the track assignment formulation. Therefore, in the proposed approach, instTerm routing is addressed with track assignment and we then use attention-based RL model to solve the most critical part in actual routing the assigned instTerms.

\subsection{Attention-based REINFORCE}
Recent works \cite{kool2018attention} on using attention-model-based REINFORCE - a reinforcement learning (RL) algorithm to solve combinatorial problems have demonstrated near optimum performance with significant generalization capability compared to existing heuristic based method. It outperforms previous work including Pointer Network (PN) \cite{vinyals2015pointer}, actor-to-critic version of PN \cite{bello2016neural} and LSTM version of PN \cite{nazari2018reinforcement} in widely studied problem sets including Travelling Salesman Problem (TSP), Orienteering Problem (OP) and price collecting TSP. In solving these problems, the solution can always be formulated as a sequential decision  process. One important reason they tend to be solved reasonably well with reinforcement learning (RL)  is that  they can be modelled as Markov Decision Process (MDP).

In solving such combinatorial problems, the attention-model-based REINFORCE use an existing policy gradient RL model: REINFORCE. In a policy gradient RL algorithm, a model is used to learn a policy model $p(a|s,\theta)$, matching state $s$ of a problem at each time step to a corresponding probability distribution of all actions $a$ by iteratively optimizing the policy model parameters over training samples. The cost that training process aims to minimize is the expectation of reward $r$ collected after certain policy $p$ has been rolled out for an episode, which  can be expressed as $E_{p}[r(\tau)]$ \cite{sutton2018reinforcement}. Following the policy gradient theorem, the gradient of the cost can be expressed as shown in Eqn.~\ref{policygradienttheo}: 

\begin{equation}
\label{policygradienttheo}
    \nabla E_{\pi_{\theta}}[r(\tau)] = E_{\pi_{\theta}}[r(\tau)(\sum_{t=1}^T \nabla log p_{\theta}(a_t|s_t))]
\end{equation}

\noindent which can be sampled and approximated from training data.

In REINFORCE, the above gradient of cost function is used to optimize or train the policy model. However, the training process of REINFORCE tends to be unstable due to the delayed reward mechanism of REINFORCE. Thus, REINFORCE with baseline is applied to stabilize REINFORCE.

In  attention-model-based REINFORCE  \cite{kool2018attention}, the formulation of problem, taking Travelling Salesman Problem (TSP) as an example, can be described as follows: the solution is defined as a tour  $\pmb{\pi} = (\pi_1, ..., \pi_n)$, which is  a sequence of the $n$ nodes (cities) in a TSP problem $s$. The input of the policy model is based on the graph structure (layout of cities) of the TSP, and the policy model output a probability distribution $p_\theta(\pi_t|s,\pmb{\pi}_{1:t-1})$ over all the nodes that are likely to be visited at the next time step $n$, during which nodes already visited are masked to ensure zero probability to be visited. Based on this, a problem policy $p(\pmb{\pi}|s)$ is defined as Eqn.~\ref{problempolicy}, which is the product of probability distribution for the $n$ steps. If at each time step, the node with the highest probability is chosen, a solution (path) is said to be given in a deterministic greedy rollout manner:  

\begin{equation}
\label{problempolicy}
    p_\theta(\pmb{\pi}|s) = \prod_{t=1}^{n} p_\theta(\pi_t|s, \pmb{\pi}_{1:t-1}) 
\end{equation}

Based on the problem policy and policy gradient theorem, the gradient of the loss function used in the attention model  is defined as Eqn.~\ref{lossgrad}. For a TSP problem, the loss term $L(\pmb{\pi})$ is the total tour length of the TSP problem instance. When applying the attention-based model to solve detailed routing problems, the loss is changed accordingly: 

\begin{equation}
\label{lossgrad}
    \nabla L(\pmb{\theta}|s) = E_{p_{\theta}(\pmb{\pi}|s)}[(L(\pmb{\pi})-b(s))\nabla log p_{\theta}(\pmb(\pi)|s)]
\end{equation}

A rollout baseline $b(s)$ is applied that is periodically updated in the following ways: $b(s)$ is the cost of a solution from a deterministic greedy rollout of the policy defined by the best model so far. In actual implementation, a $t$-test is applied to ensure that the baseline is based on the best model so far. 

\begin{figure}[thpb]
\centering
\includegraphics[width=0.45\textwidth]{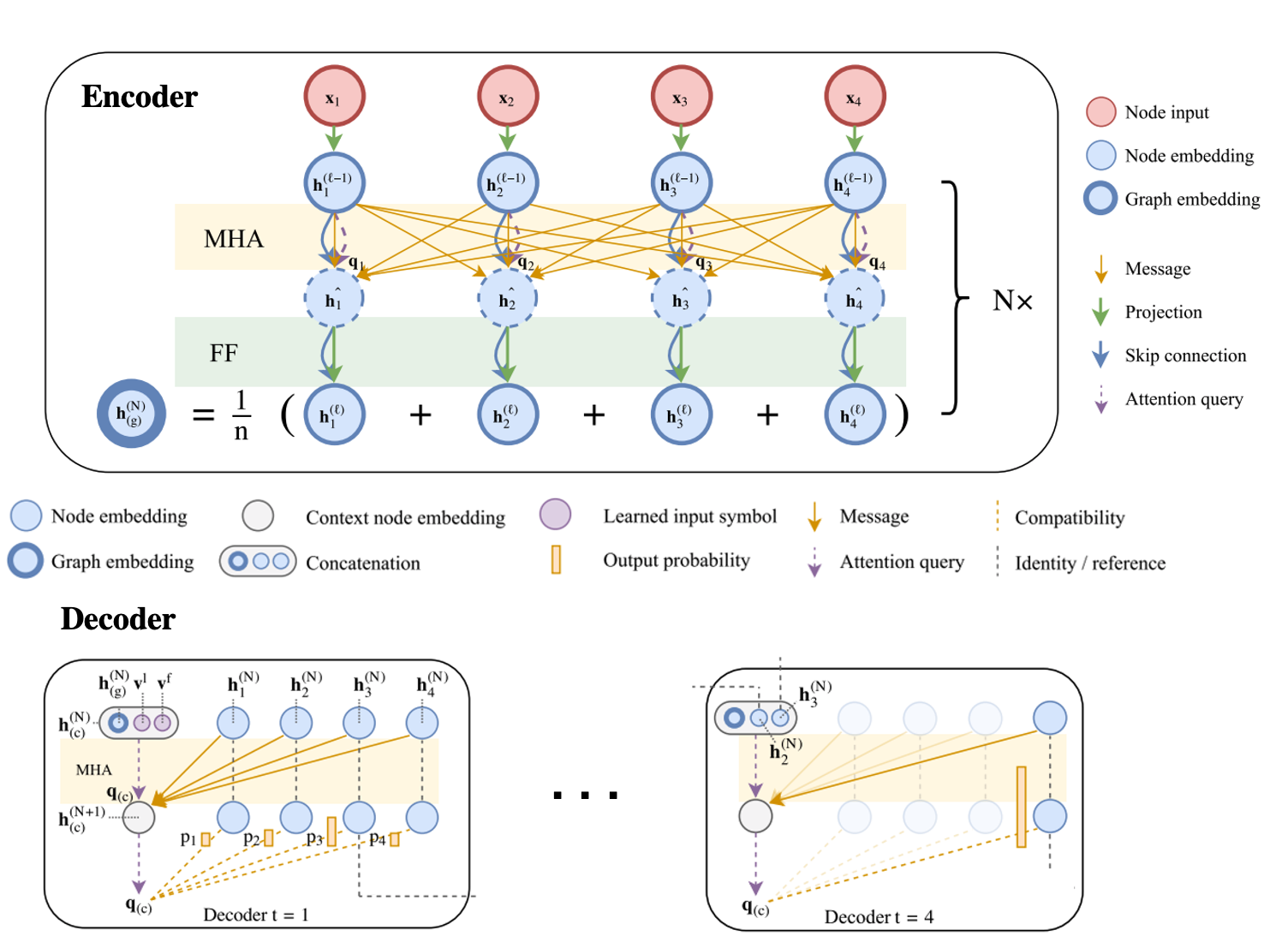}
\caption{Encoder-decoder model structures. Adopted from \cite{kool2018attention}.}
\label{encoderdecoder}
\end{figure}

The policy model is realized with attention-based encoder-decoder model which can be considered as a Graph Attention Network \cite{velivckovic2017graph}, as shown in Fig.~\ref{encoderdecoder}. The encoder is similar to the Transformer architecture \cite{vaswani2017attention} encoder. After a first learned linear projection layer, the major part of encoder consists of N attention layers, with each layer consisting of two sublayers: a multi-head attention (MHA) layer and a fully connected feed-forward (FF) layer, skip-connection \cite{he2016deep} and batch normalization (BN) \cite{ioffe2015batch} are applied at each sublayer. Formally, the attention layers can be expressed as follows:

\begin{equation}
\label{mha1}
    \hat{h_i} = BN^l (h_i^{(l-1)} + MHA_i^l(h_1^{(l-1)},...,h_n^{(l-1)}))
\end{equation}

\begin{equation}
\label{mha2}
    h_i^{(l)} = BN^l (\hat{h_i} + FF^l(\hat{h_i}))
\end{equation}

\noindent where $\hat{h_i^{(l)}}$ represents the output values (in vector form) in the $i$ th node of $l$ th MHA layer and $h_i$ represents the output values of the $\hat{h_i}^{(l)}$ after BN.

At the heart of the MHA structure is the attention mechanism which can be summarized as a weighted message passing between the nodes in a graph. The weight of a message \textit{value} that a node receives from a neighbor depends on the \textit{compatibility} of its \textit{query} with the \textit{key} of the neighbor \cite{kool2018attention}. For each node, its corresponding \textit{key, query} and \textit{value} is obtained by projecting the node embedding/input by parameter matrices correspondingly, whose weights are automatically learned during training. The decoder consists of only one layer of MHA and it outputs the probability distribution of nodes to be visited at each time step based on the embeddings from the encoder and the output generated at the previous time steps. 

Inspired by the similarity between existing combinatorial problems and the critical sequencing step underlying detailed routing algorithms, we propose an attention-based REINFORCE model as a new way to address the detailed routing problem. In applying the attention-based model to solve detailed routing, while the model architecture remains the same, the loss function is modified and the definition of a node now becomes a pair of instTerms, which will be addressed in details Method. (Another motivation for applying learning-based algorithm to solve detailed routing is the difficulty to apply a robust heuristics based or hard-coded method to the sequencing step. It is worth mentioning that although there exists simpler algorithms \cite{kool2018attention} for standard combinatorial problems such as Nearest Neighbors for TSP , the unique hierarchical nature and complexity of IC physical design flow and routing makes these algorithms not readily applicable to solve detailed routing\cite{liao2020deep}.)


\subsection{Genetic Algorithm}
\begin{figure}[h]
\includegraphics[width=0.45\textwidth]{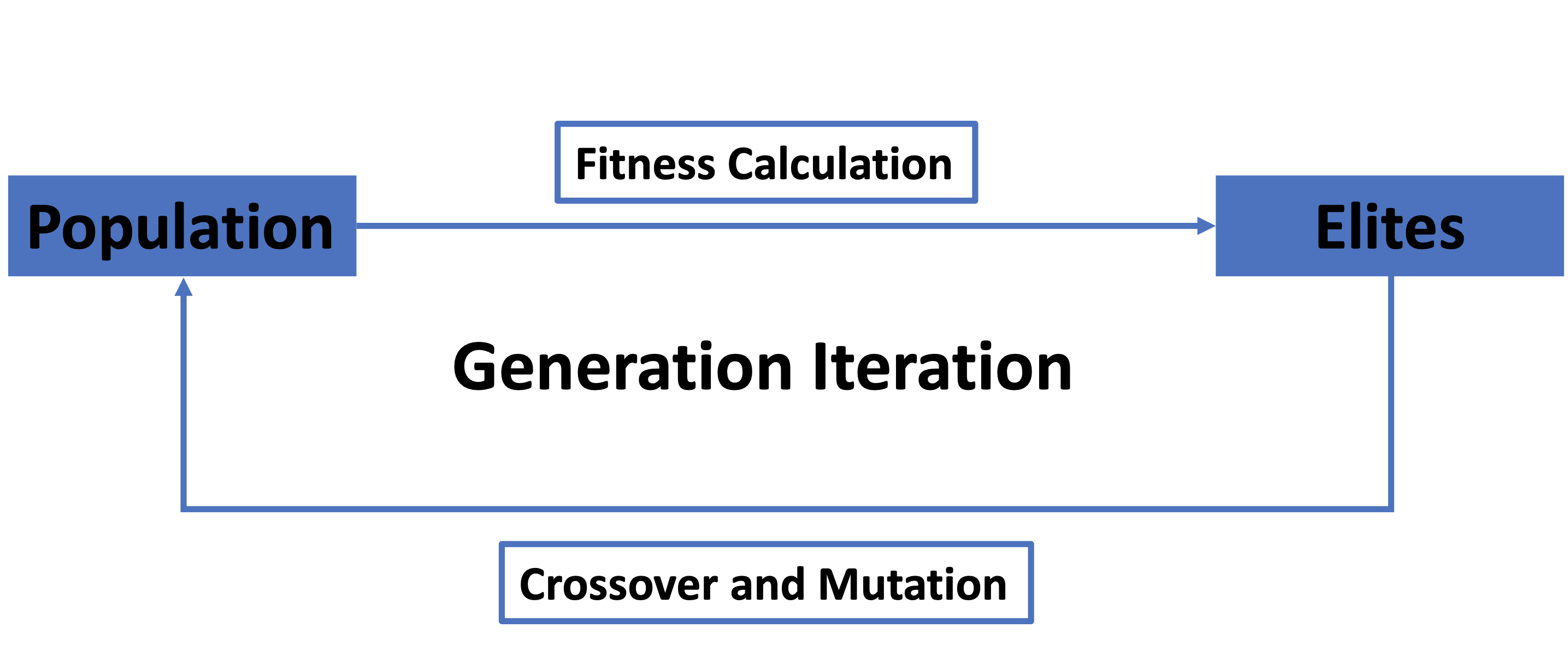}
\caption{Generation iteration of genetic algorithm (GA).}
\label{GA} 
\end{figure}

Genetic algorithms (GA) \cite{whitley1994genetic}  have been widely used to solve combinatorial optimization problems \cite{jing2003parallel, muhlenbein1992parallel}. It is realized with iterations of generations, as schematically illustrated in Fig.~\ref{GA}. For one generation, a \textit{population} consisting  a pool of chromosome is firstly generated either randomly or from elite parents of previous generation. A \textit{fitness} function is then applied to calculate the fitness of each chromosome in the population, a proportion of the chromosome in the original population that has higher fitness scores are selected to be the \textit{elites}, which naturally becomes parents for generating the next generation of population. A new generation of population is generated by \textit{crossover} and \textit{mutation} operations of chromosome among elites.
In the next generation iteration, the previous population is replaced by newly generated population.

In this work, GA algorithm, as a comparison to attention model, is used in genetic router for determining the sequence of instTerm pairs to be routed, which significantly determines the quality of detailed routing solutions. GA has been one of the best methods in solving combinatorial optimization problems in IC physical design \cite{lienig1993genetic,lienig1996parallel, esbensen1994macro}, especially when no other heuristics and learning methods are not readily available. Unfortunately, although it tends to work well in small scale problems with no stringent run time requirements, it suffers from lack of generalization ability, large run time cost and limited scalability. In this work, we propose to use attention router as an alternative to the genetic router in track-assignment detailed routing.  
\\

\section{METHOD}
\begin{figure}[h]
\includegraphics[width=0.45\textwidth]{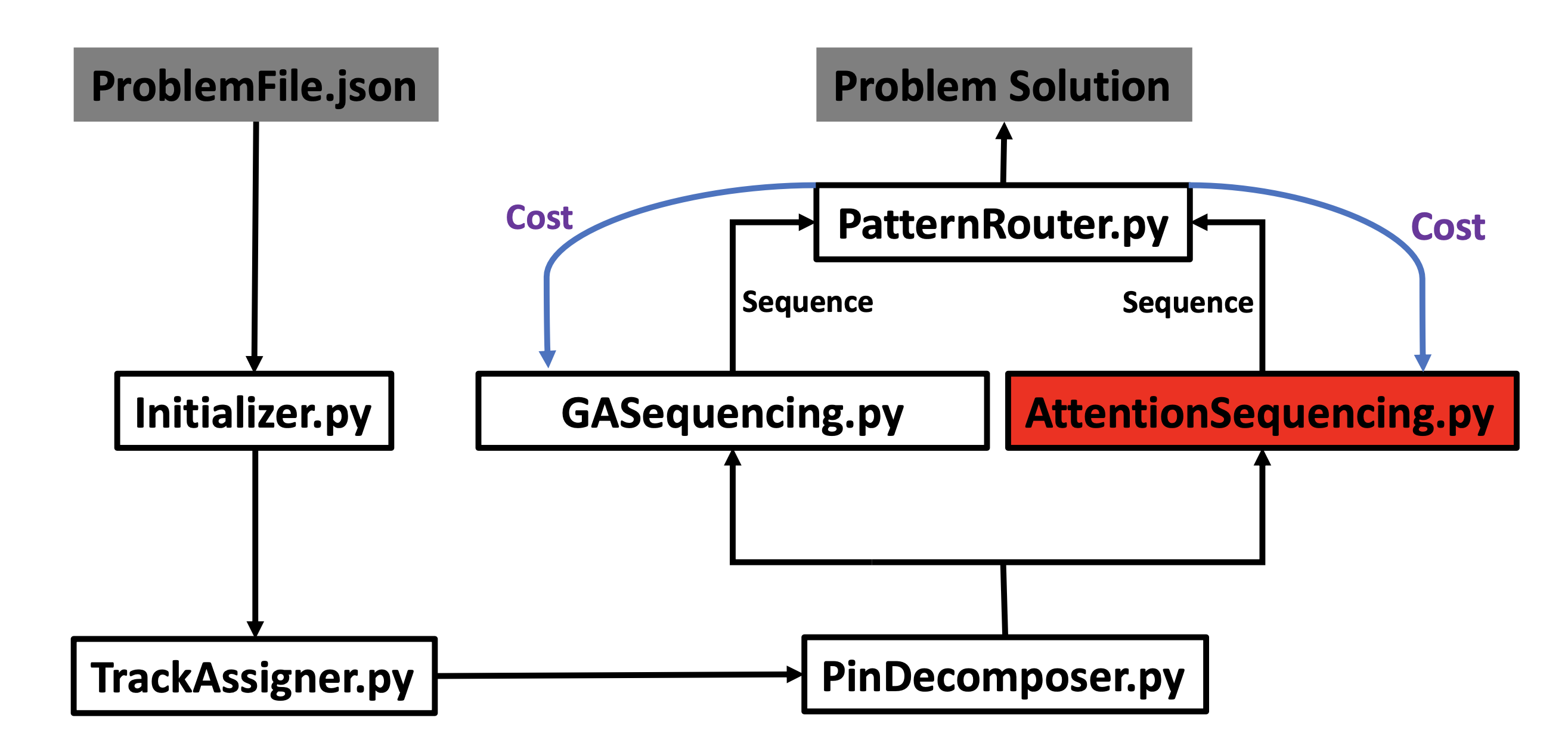}
\caption{Pipeline of our attention routing.}
\label{pipeline} 
\end{figure}

Fig.~\ref{pipeline} illustrates our attention router model. Firstly, all problem files in a specific problem set describing the design information including instTerms locations and nets information is read in and parsed by the \textit{Initializer}. \textit{Track Assigner} is then applied to complete the track assignments for all instTerms in each problem. Once complete, the exact locations of all instTerms are determined. Next, \textit{Pin Decomposer} is then applied to all the nets of each problem to further simplify each problem for the subsequent routing in the form of a set of instTerm pairs.

In routing instTerm pairs, GA (for genetic router) and attention-based REINFORCE model (for attention router) are applied to determine the sequence of the instTerm pairs in a problem to be routed. \textit{GA Sequencing} is executed only once, while \textit{Attention Sequencing} is firstly trained by solving problems from the training set, and then applied to problems in both training and test sets. During execution of \textit{GA Sequencing} and \textit{Attention Sequencing}, the same \textit {Pattern Router} function is utilized to compute the exact routes connecting each instTerm pairs and to calculate the cost of the solution. The problem solution is a concatenation of the actual routes for all the instTerm pairs and the cost of a solution is defined as a weighted sum of \textit{wirelength} (WL) and \textit{number of openings} ($\#$Open), which is the number of instTerms pairs that remain unconnected due to a lack of feasible route for the given problem. The cost is given in Eqn.~\ref{cost}. Since openings is highly undesirable in the physical design process, in this research weights are set as: $w_1 = 1, w_2 = 10$

\begin{equation}
\label{cost}
    Cost = w_1 * WL + w_2 * \#Open
\end{equation}

The details of the individual modules of our method are provided next. Python and Pytorch (Machine Learning Framework) are used for implementing the proposed algorithm. 

\subsection{Track Assigner}
\label{track_assignment_section}
The first step of routing is to assign instTerms to WSP tracks. As the x-coordinates of all instTerms are fixed, the routing problem is reduced to finding the appropriate track while satisfying the assignment rules and ensuring no short circuits. 

\begin{figure*}[thpb]
\centering
\includegraphics[width=0.85\textwidth]{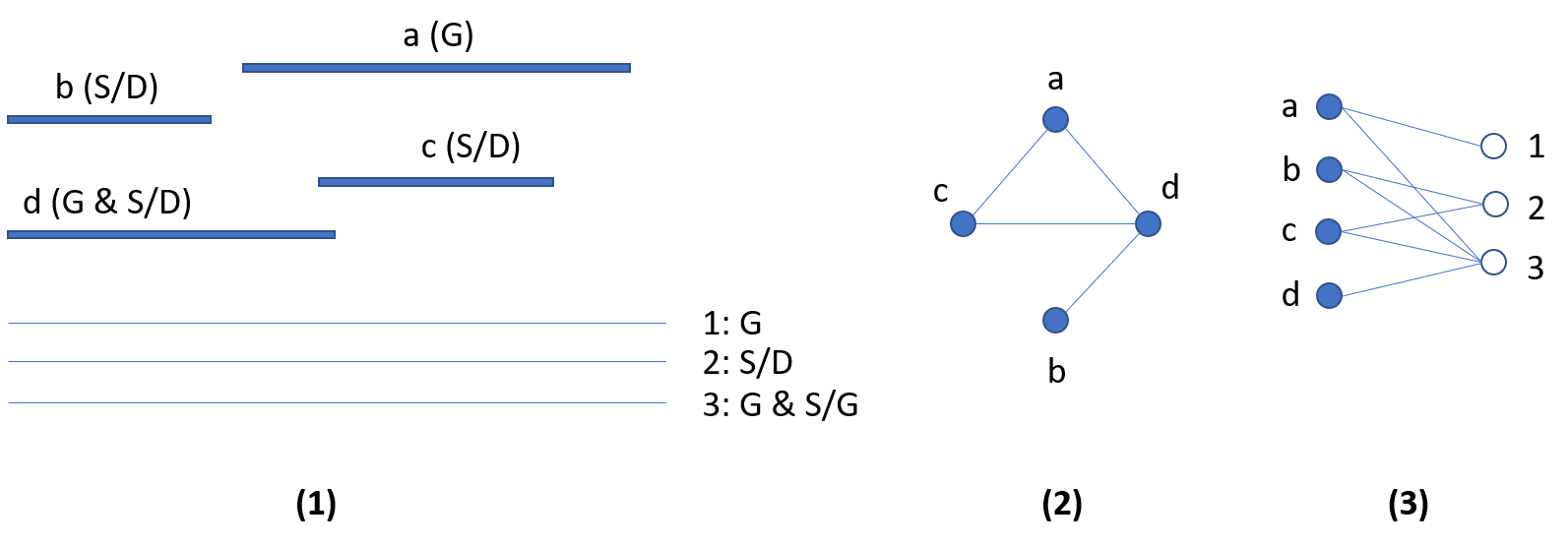}
\caption{Illustration of the track assignment problem: (1) the instTerms and tracks, (2) the overlap graph, (3) the assignment graph.}
\label{track_assign_illustration} 
\end{figure*}

Not all tracks can be used to route an instTerm. For illustration, in Fig~\ref{track_assign_illustration} (a) with three tracks, Track 1 and 2 can only contain gate (G) and source/drain (S/D) terminals, while Track 3 can have both G and S/D terminals. Specifically, in this work, we have seven tracks per row, where the G terminals should be on tracks 1, 2, 6, and 7, S/D terminals on tracks 2, 3, 4, 5, and 6, hence for instTerms composed of both G and S/D, only track 2 and 6 shall be used.

Two graphs are used in the track assigner, namely the \textit{overlap graph} and the \textit{assignment graph} (Fig~\ref{track_assign_illustration}). The overlap graph models the conflicts between instTerms, where each node represents an instTerm, and an edge exists between two nodes if they belong to different nets and their x-ranges overlap (implying that they cannot be assigned to the same track). This constitutes a horizontal constraint graph \cite{sherwani2012algorithms}. The assignment graph is a weighted bipartite graph, the nodes on one side are the instTerms and the other are the available tracks, if an instTerm is assignable to a track, these two nodes are connected by an edge with a weight as the assignment cost. Because instTerm is constrained to route on the tracks of the containing row, the vertical connection and via costs can be omitted, and we model the cost considering only the horizontal track utilization.

With the help of the two graphs, the track assignment problem is reduced to matching the instTerm nodes to the track nodes in the assignment graph while minimizing the matching cost, such that no conflicting instTerm nodes in the overlap graph are matched to the same track. For the standard bipartite matching problem, \cite{karp1990optimal} provides a polynomial algorithm, but the problem now becomes NP-complete \cite{batterywala2002track} after introducing the assignment conflict constraint. As instTerm splitting is not allowed, we modified the algorithm presented in \cite{batterywala2002track} to solve the instTerm track assignment problem, and the algorithm is described in Algorithm~\ref{ta_algo}.

\begin{algorithm}
\SetAlgoLined
\SetKwInOut{Input}{Input}\SetKwInOut{Output}{Output}
\Input{Netlist containing the instTerm and track information}
\Output{Assigned instTerm-track pairs}
 Build overlap graph $G_O = (V_o, E_0)$ and assignment graph $G_A = (V_A, E_A, w)$\;
 \While{Exists assignable instTerms}{
  Find the largest clique $K_m$ in $G_O$\;
  Perform weighted bipartite matching on the sub-graph $G_m = (V_m, E_m)$, where $V_m \subseteq V_A, E_m \subseteq E_A, V_m \in K_m$\;
  Assign the uniquely assignable instTerms to the corresponding track (look-ahead heuristic in \cite{batterywala2002track})\;
  Update $G_O$ and $G_A$: remove assigned instTerm nodes and associated edges\;
 }
 \caption{The track assignment algorithm.}
 \label{ta_algo}
\end{algorithm}

\subsection{Pin Decomposer}
\begin{figure}[htbp]
\centering
\includegraphics[width=0.35\textwidth]{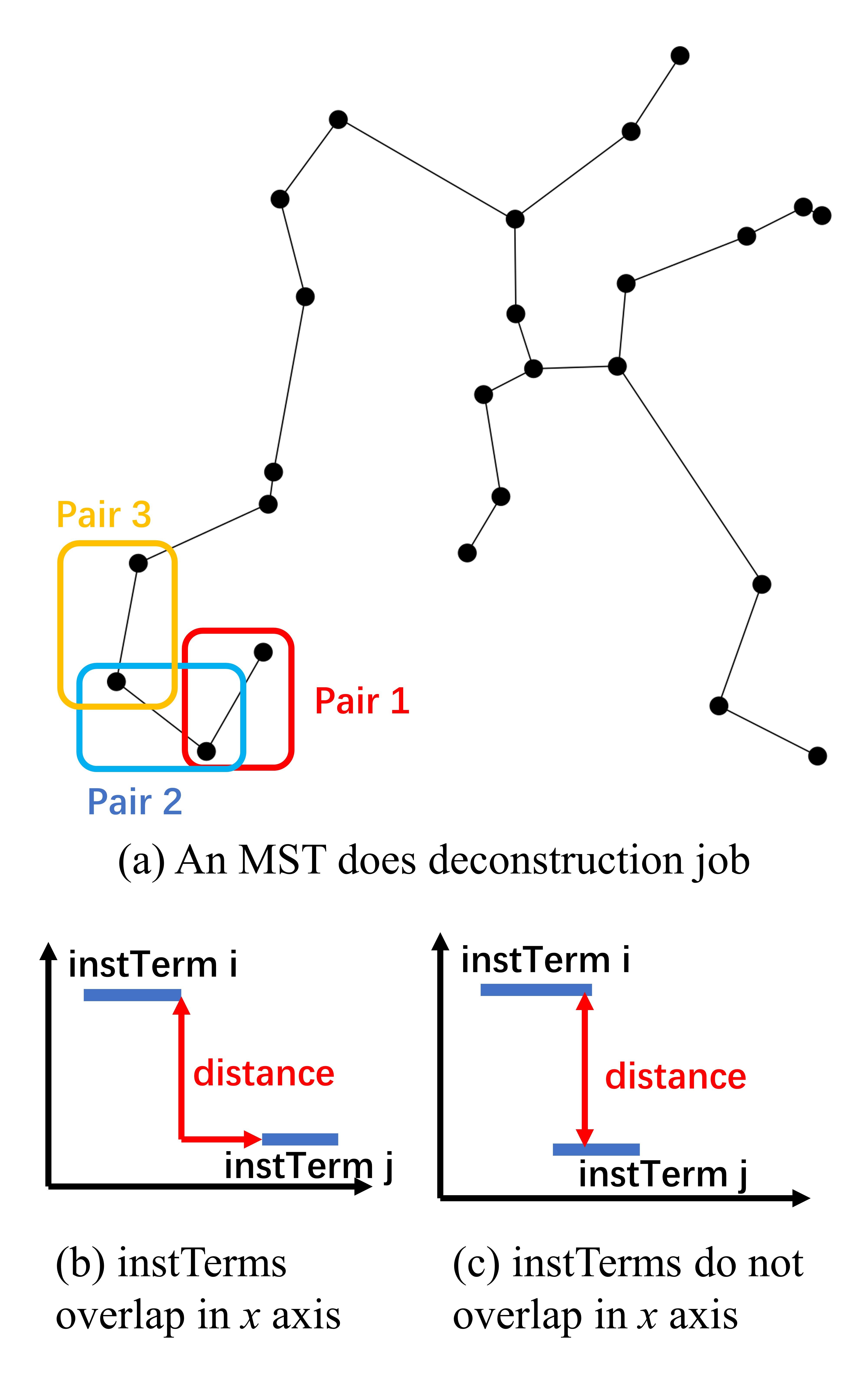}
\caption{Pin decomposer. (a) An MST reveals in the instTerm pairs. (b)(c) Calculation of the distance between two instTerms.}
\label{Pin Decomposer} 
\end{figure}

Each net is composed of multiple instTerms. Each instTerm has the coordinate of ($x_{1}$, $x_{2}$, $y$) $, x_1,x_2,y \in \mathbf{Z^+}$. In order to simplify the problem, instead of directly working on the sequence of nets, we first decompose each net into multiple two-instTerm pairs, so that after the decomposition, our model will produce the best sequence of these instTerm pairs. Kruskal's algorithm \cite{kruskal1956shortest} is utilized to construct a Minimum Spanning Tree (MST) first, as the MST naturally reveals the pin pairs that should be connected as shown  in Fig.~\ref{Pin Decomposer}a.


In order to create the MST, a distance matrix is needed, where each element $(i,j)$ in this matrix the distance between instTerm $i$ and instTerm $j$. However, due to the fact that we are dealing with instTerms (bars) instead of nodes (points), distance is computed as the minimum Manhattan distance between the instTerm bars as shown in Fig.~\ref{Pin Decomposer}b and Fig.~\ref{Pin Decomposer}c. Note that even after this decomposition, we are still dealing with instTerm pairs rather than node pairs.

\subsection{Pattern Router}
We route each instTerm pair sequentially using a simplified pattern router \cite{kastner2002pattern} in the rectilinear space that can be modelled as a graph $G(V,E)$. In the graph, each edge has a capacity $c_{ij}$, which is initialized to 1 before routing. In routing the instTerm pairs of a problem, the routable paths are edges $e_{ij}$, $i,j \in \mathbf{Z^+}$ with non-zero capacity.

\subsubsection{Routing two vertices}
We loop through all combinations of ($v_{i}$, $v_{j}$), where $v_{i}$ is a vertex from instTerm $i$, and $v_{j}$ is a vertex from instTerm $j$. For each combination, we use our simplified pattern router, where we only consider ``L'' patterns and then ``Z'' patterns if ``L'' patterns fail.

\begin{figure}[htbp]
\centering
\includegraphics[width=0.46\textwidth]{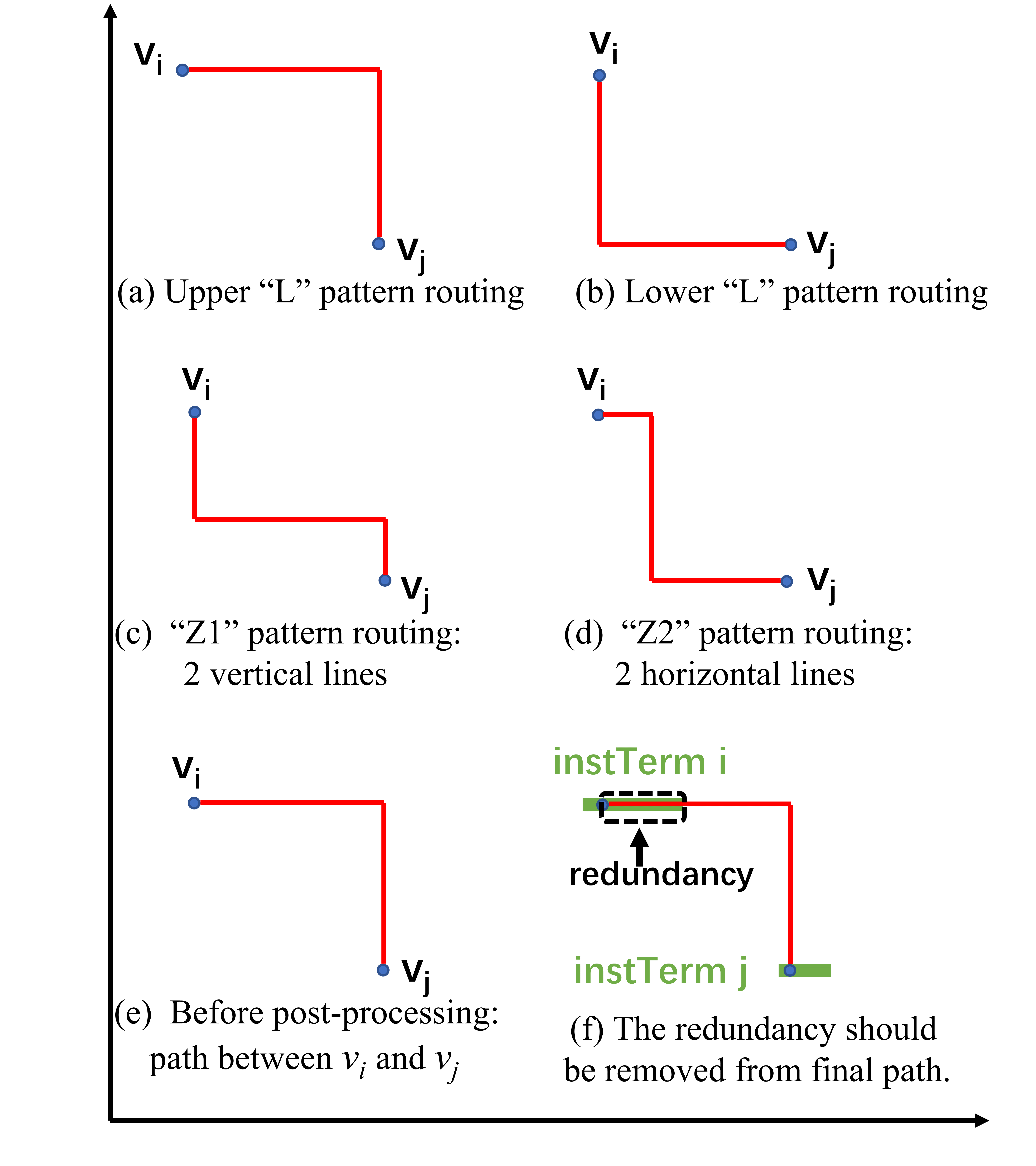}
\caption{Two pin router. (a)(b) ``L'' pattern routing, (c)(d) ``Z'' pattern routing, (e)(f) Post Processing.}
\label{Two Pin Router} 
\end{figure}

\textbf{``L'' pattern routing} There are 2 kinds of ``L'' patterns: upper ``L'' and lower ``L'', which are shown in Fig.~\ref{Two Pin Router}a and Fig.~\ref{Two Pin Router}b respectively. Straight lines are also considered as a special case of  ``L'' pattern, when two instTerms overlap in the $x$ axis or share the same $y$ coordinate.

\textbf{``Z'' pattern routing} If ``L'' patterns fail, our router employs ``Z'' pattern routing. There are also two kinds of ``Z'' patterns, as shown in Fig.~\ref{Two Pin Router}c and Fig.~\ref{Two Pin Router}d.

If no above patterns can route  ($v_{i}$, $v_{j}$), an opening occurs.

\subsubsection{Post processing}
Once $v_{i}$ and $v_{j}$ are routed successfully, we obtain the path that connects them. Any redundancies in the path are  removed. Fig.~\ref{Two Pin Router}e and Fig.~\ref{Two Pin Router}f illustrate that the redundancy is the overlapping parts between the path and instTerms $i$, $j$.

\subsection{Attention Model Implementation}
In order to find an optimized routing solution that minimizes the loss among all possible routing sequences, we use an attention based encoder-decoder model with a rollout baseline. We define each problem instance as a graph with $n$ nodes, and each node $n_i, i \in {1,...,n}$ is represented by an instTerm pair between instTerm $i$ and instTerm $j$. Each instTerm pair is in the form:

$(x_{i1}, x_{i2}, y_i, x_{j1}, x_{j2}, y_j, l)$, $\forall i, j, i \neq j$, 

\noindent where $x_{i1}, x_{i2}, y_i$ represents the xy-coordinates of instTerm $i$. Similarly, $x_{j1}, x_{j2}, y_j$ represents the xy-coordinate of instTerm $j$; $l$ represents the net index which is later used in the routing part and thus will not be discussed in detail here. We define the solution to the routing sequence $\pi$ as a permutation of the $n$ nodes. 

Since the number of instTerms varies in each routing problem, to ensure that each problem instance $s$ has the same number of nodes $n$ (inst pairs), we perform two possible padding strategies on the problem instance: Pad Random and Pad Empty. In the Pad Random strategy, we uniformly sample xy-coordinates in the domain of all instTerms; in the Pad Empty strategy, we are not concerned with the actual coordinates and instead pad with all-zero nodes in the form of $(0, 0, 0, 0, 0, 0, 0)$. In each strategy, we set the graph size $n$ to be the maximum graph size of all the problem instances, and pad nodes to problem instances of smaller size. After experiments with the two padding strategies, we decide to use the Pad Empty strategy, which performs more stably and generates routing sequences with smaller loss. We think the result of the Pad Random strategy can be improved if the coordinate sampling is done based on the original coordinate distribution of the instTerms instead of a uniform sampling.

After careful parameter tuning, we set the number of training batches $B = 20$, and for each batch, the training batch size $T = 5$. We train our model on epoch sizes $E = 100$. 

Our current problem sets contains different total number of problem instances, which we split into 60\% as training, 20\% as validation, and 20\% as test cases. Each epoch is a walkthrough of all the problem instances in the training set. In each batch, the dataset loader loads 5 out of the training problem instances in sequence as the current training batch. At the end of each epoch, the model is evaluated on the validation set, and the average loss is computed. After completion of all epochs, the model with the smallest average loss is used to evaluate the test set by generating corresponding routing sequences to each  test problem. 

We define the loss $L(\theta|s) = E_{p_{\theta}(\pi|s)}[L(\pi)]$, where $L(\pi)$ is a vector of length 5, containing losses of each problem instance returned by the pattern router as discussed in section 3.4. The loss of each problem instance returned by the pattern router is defined as a weighted sum of the total wire length and number of openings using the routing sequence $\pi$ generated by the attention model, as shown in Eqn.~\ref{cost}. We optimize the loss $L(\theta|s)$ by gradient descent, using the REINFORCE gradient estimator with rollout baseline $b(s)$ \cite{kool2018attention}:

\begin{equation}
\nabla L(\boldsymbol{\theta} | s)=\mathbf{E}_{p_{\boldsymbol{\theta}}(\boldsymbol{\pi} | s)}\left[(L(\boldsymbol{\pi})-b(s)) \nabla \log p_{\boldsymbol{\theta}}(\boldsymbol{\pi} | s)\right]
\end{equation}

At the end of each epoch, we perform a one-sided t-test between the current model and the baseline model with a significance parameter $\alpha$ to decide whether or not the baseline model should be updated. The algorithm is described in Algorithm~\ref{ta_algo}.

\begin{algorithm}
\caption{Attention Sequencing}
\label{ta_algo}
\SetAlgoLined
\SetKwInOut{Input}{Input}\SetKwInOut{Output}{Output}
\Input{Number of epochs E, batch size B, training set T, significance $\alpha$}
\Output{Sequence based on best policy}
Init $\theta$, $\theta^{BL}$ $\leftarrow$ $\theta$ \;
\For{epoch=1,...,$E$}{
 \For{batch=1,...,$B$}{
    $t_i$ $\leftarrow$ SampleInstance() $\forall i \in {1,...,T}$\;
    $\pi_i$ $\leftarrow$ SampleRollout($t_i$, $p_{\theta}$) $\forall i \in {1,...,T}$\; 
    $\pi_i^{BL}$ $\leftarrow$ GreedyRollout($t_i$, $p_{\theta^{BL}}$) $\forall i \in {1,...,T}$\;
    $\Delta L$ $\leftarrow$ $\sum_{i=1}^B{
    (L(\pi_i)-L(\pi_i^{BL}))\Delta_{\theta}\log{p_{\theta}(\pi_i)}}$\;
    $\theta$ $\leftarrow$ Adam($\theta$, $\Delta L$)\;
 }
 \If{OneSidedPairedTTest($p_{\theta}$, $p_{\theta^{BL}}$) $\leq$ $\alpha$}{
    $\theta^{BL}$ $\leftarrow$ $\theta$\;
 } 
}
\end{algorithm}

\subsection{Genetic Algorithm (GA) Sequencing}

The GA-based sequencing, which works as a comparison to attention-based model in this work follows the typical generation iterations of the GA algorithm shown in Fig.~\ref{GA}, with crossover and mutation operations within each generation. The details of the GA sequencing are shown in Algorithm~\ref{ga_algo}. In this problem, each chromosome consists of an ordered vector of numbers, representing the routing sequence for all instTerms pairs in a problem. Since we are trying to minimize the cost in Eqn.~\ref{cost}, the fitness  of a chromosome is the negative value of the cost in Eqn.~\ref{cost} by solving the corresponding problem with the sequence indicated by the chromosome. Model parameters for the GA sequencing are set as: \textit{generation number}: 10, \textit{population size}: 10, \textit{elites size}: 4, \textit{number of mutations}: 1. Note  that a limited number of generations  is chosen to avoid the run time of GA sequencing from becoming too long. 

\begin{figure}[h]
\includegraphics[width=0.45\textwidth]{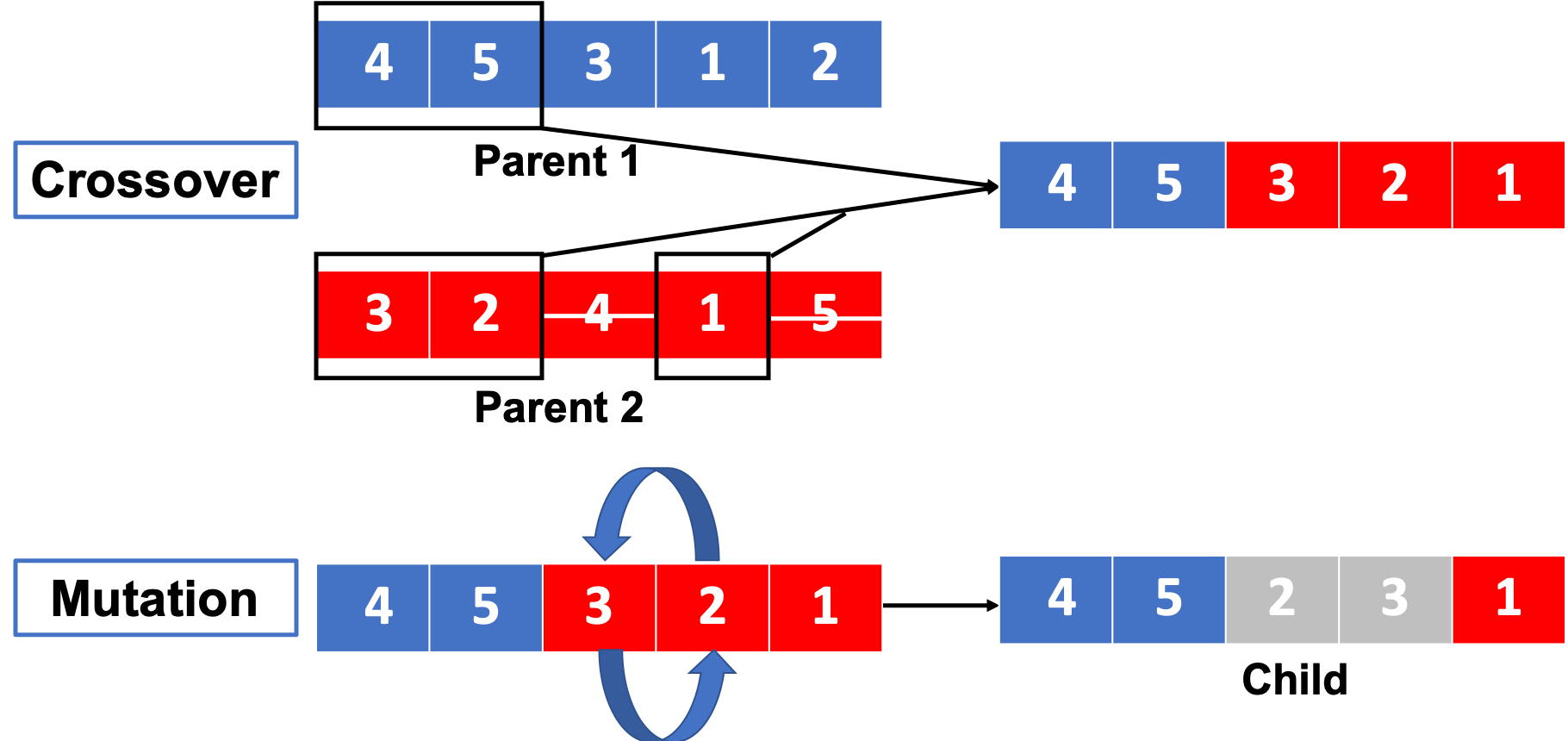}
\caption{Crossover and mutation methods used in the GA sequencing.}
\label{cross_mutation} 
\end{figure}

Since each chromosome in our algorithm is a sequence rather than independent numbers, each number  can only appear once a chromosome. To address this uniqueness, the crossover and mutation operations adopted in research is demonstrated in Fig.~\ref{cross_mutation}.  In generating a new child's chromosome, partially matched crossover is adopted. After crossover, a newly generated chromosome is obtained. In the mutation step, two genes in two random selected locations in the newly generated chromosome switch their positions. This crossover and mutation method guarantee that all generated kids represent a legal sequence. 

\begin{algorithm}
\caption{Genetic Algorithm Sequencing}
\label{ga_algo}
\SetAlgoLined
\SetKwInOut{Input}{Input}\SetKwInOut{Output}{Output}
\Input{Number of generations G, population size P, elites size Q, number of mutations M}
\Output{Last generation of sequencing (chromosome)}
Init chromosomes \{$C_1, ..., C_P$\} in first generation  \;
\For{generation=1,...,$G$}{
 Select elites {$E_1$,..., $E_Q$} based on Fitness Score\;
 \For{i=1,...,P}{
 $C_i$ $\leftarrow$ CrossOver($E_i$, $E_j$) $i,j \in 1,...,Q$\;
 $C_i$ $\leftarrow$ Mutation($C_i$)\;
 }
 } 
\end{algorithm}

\section{EXPERIMENTS}
In order to assess the performance of the attention router and its comparison to the baseline genetic router, both algorithms are applied to detailed routing problems from two problem sets: \textit{Small} and \textit{Large}, which are both analog design problems based on commercial advanced node technologies (sub-16 nm technology). To be specific, \textit{Small} problem set consists of different placement solutions for Comparators and OpAmp, while \textit{Large} problem set consists of different placement solutions of Analog-to-Digital Converter (ADC). For \textit{Small} problem set, the number of instTerms for each problem range from 10 to 100, and in \textit{Large} problem sets, the number of instTerms for each problem range from 100 to 1000. 

We ran experiments based on the two problem sets. \textcolor{black}{Three experiments were conducted using the \textit{Small} data set with 100, 500, and 5000 training problems denoted as \textit{Small100}, \textit{Small500}, and \textit{Small5000} respectively. Two experiments were conducted using the \textit{Large} data set with 100 and 500 training problems denoted as \textit{Large100} and  \textit{Large500} respectively.} In the genetic router, GA Sequencing is run for each of the problems in the problem sets. In the attention router, attention sequencing is trained iteratively using the training sets and then applied to previously unseen problems in the test sets. For the four sets of experiments, the key  parameters for the attention model are: \textit{batch size}= 5 and \textit{epoch number}= 100. Increasing the batch size and the number of epochs   significantly improves the attention model's performance, (thus we set the epoch number to allow the model gain enough learning experience, while not spending too much time for training.). All experiments are run on a workstation with an Intel Core i7-6850 CPU \textcolor{black}{}. In training the attention router, it takes around 6 minutes for a training epoch on problem sets $Small500$ and around 25 minutes for a training epoch on problem sets $Large500$ .

\section{RESULTS AND DISCUSSIONS}
\subsection{Training}
\begin{figure}[h]
\includegraphics[width=0.5\textwidth]{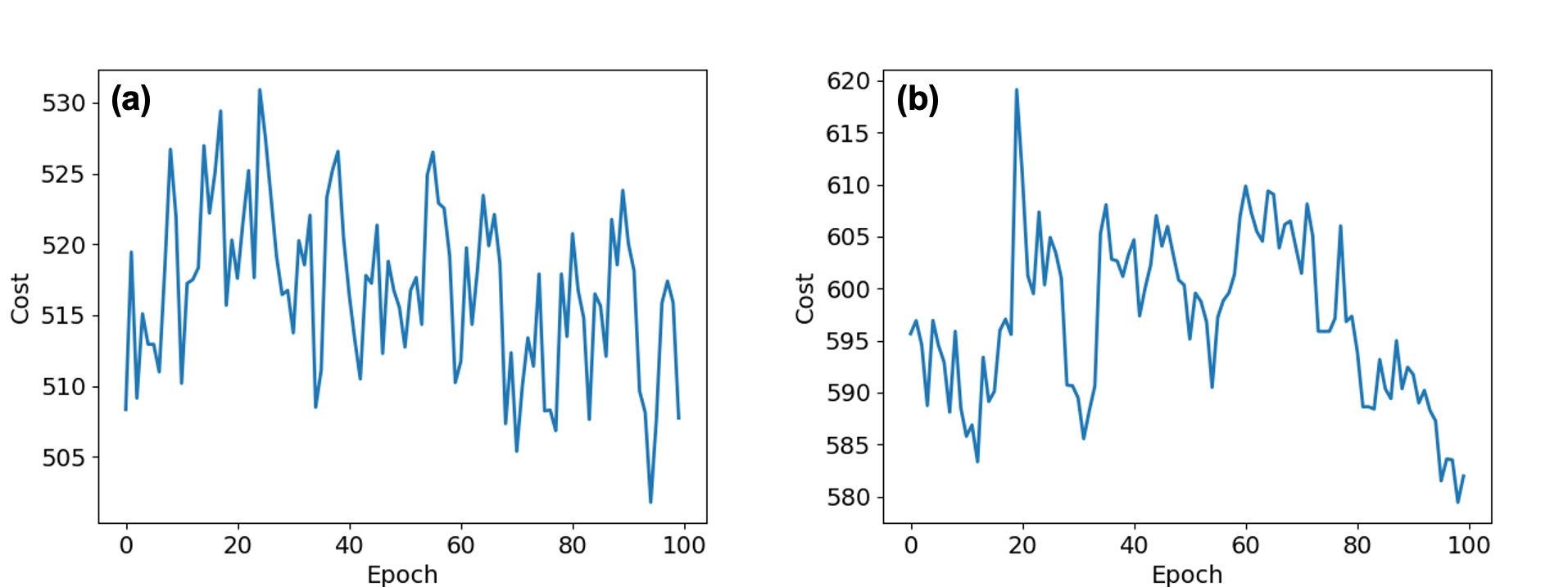}
\caption{Cost vs. training epochs on (a) \textit{Small100}, and (b) \textit{Small500} sets. }
\label{train_curve} 
\end{figure}

Fig.~\ref{train_curve} shows  cost versus training epochs plots for problem sets \textit{Small100} and \textit{Small500}.  The high variation during the training process is an intrinsic property of the REINFORCE policy gradient algorithm  used in the attention model. It can be explained by the ``delayed reward'' of policy gradient REINFORCE algorithm used to optimize the network, as shown in Eqn.~\ref{policygradienttheo}. In the equation, the reward signal is not obtained until $T$ steps of actions $\{a_1, ..., a_T\}$ have been taken, then it is multiplied with the summation of log-values of $p_{\theta}(a_t|s_t)$ at each step to form the gradient values for optimizing the policy networks. The delayed reward signal mechanism makes the training unstable as there is no clear guidance in terms of each action's contribution in the action sequence $\{a_1, ..., a_T\}$ to the reward. As such, the gradient based on the policy gradient theorem equation in Eqn.~\ref{policygradienttheo} can only optimize the policy networks with a rough guidance in terms of the optimization directions, instead of a more desirable one that can lead to monotonically decreasing cost values. 

The variation in the training process of REINFORCE is remedied with the introduction of a baseline \cite{sutton2018reinforcement}, which can be described as a gauge for the difficulties of problem the model is solving and usually leads to faster learning in the REINFORCE model. In this work,  although a baseline has been implemented, the variation in  training  is still present, which might be further reduced with techniques such as decay learning rate and application of critic networks \cite{kool2018attention}, which will be a part of our future work.   

\subsection{Attention router performance}

\subsubsection{Performance comparison between attention model and GA.}

\begin{figure}[h]
\includegraphics[width=0.5\textwidth]{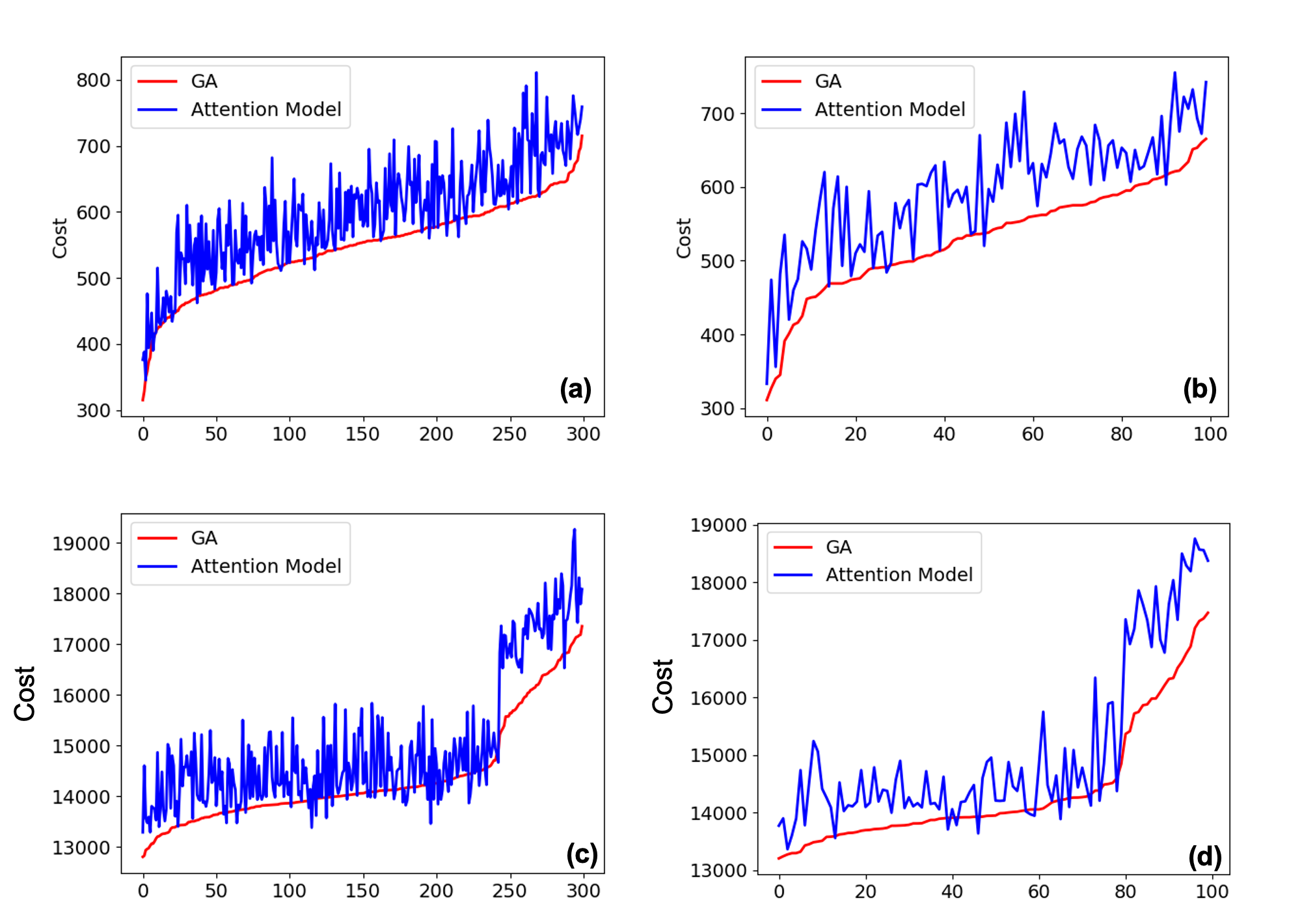}
\caption{Cost comparison across different problems: (a) \textit{Small500} training, (b) \textit{Small500} test,  (c) \textit{Large500} training, (d) \textit{Large500} test.}
\label{cost_compare} 
\end{figure}

Figure ~\ref{cost_compare} shows the cost comparison between the attention router and the genetic router for all problems in the training  and test sets for $Small500$ and $Large500$ problems. In each figure, the horizontal axis corresponds to problem indices sorted based on an ascending order of the genetic router results. For problem sets $Small500$,  while the genetic router performs better in almost all problems compared to the attention router,  the difference in cost for a given problem between the two routers is rather small (mostly within 100). For problem sets $Large500$, which has approximately ten times the number of instTerms than $Small500$ in each problem, while the genetic router still performs better overall, the number of cases in which the attention router outperforms GA is higher. It has been argued in prior work \cite{rivera2001scalable} that when applying a genetic router to solve large scale problems, it tends to exhibit high  computational cost accompanied by a degradation of the solution quality. This is primarily due to the increased complexity of problems, where the number of possible sequences of $n$ instTerm pairs is $O(n!)$. This makes increasingly larger problems intractable for GA with limited computational cost. 

By comparing the performance of the attention router on training and test sets (as shown in Fig.~\ref{cost_compare}), it can be seen that in both problem sets, the performance of the attention router is similar in training sets and test sets. This implies that the attention router can solve previously unseen problems once it is trained on the training set.  This is due to the attention model's ability to learn proper strategies across various spatial structures of the instTerm pairs. The multi-head attention (MHA) mechanism can be seen as a message passing method that allows each instTerm pair to communicate with all other instTerm pairs in the same problem regarding their relative spatial information \cite{kool2018attention}. In this way, an instTerm pair's spatial configuration within a space shared by other instTerm pairs can be continuously monitored and factored in. This spatial information is then used to form the sequential decisions for the final sequencing of the instTerm pairs. 

\begin{figure}[h]
\includegraphics[width=0.5\textwidth]{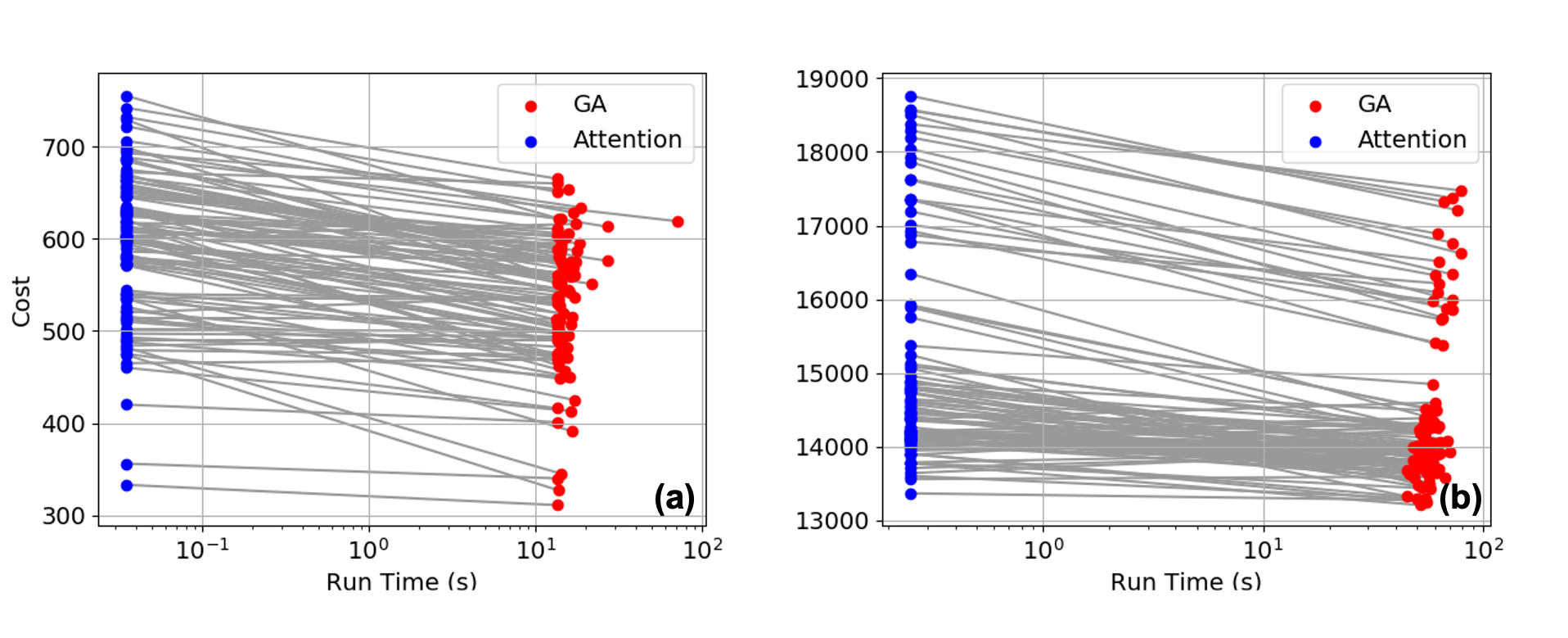}
\caption{Cost vs. run time on test sets for problems in (a) \textit{Small500}, (b) \textit{Large500}. The same problems are connected with gray lines.}
\label{costruntime} 
\end{figure}

Figure ~\ref{costruntime} compares the cost vs. run time  of the attention router and the genetic router on problem sets \textit{Small500} and \textit{Large500}. Attention router's results are shown in blue dots, while genetic router's results are shown in red dots. In both plots, while the cost range of attention router's results is slightly higher than the genetic router ones, the run time of attention router is more than two orders of magnitude ($100\times$) shorter than the genetic router: For $Small500$ problem set, genetic router takes more than 10 seconds to solve each problem, while for the attention router, the time is  less than 0.1 seconds. For the $Large500$ problem set, the genetic router takes close to 100 seconds for each problem, while the attention router only takes a little more than 0.1 seconds to solve a problem.

\begin{figure}[h]
\includegraphics[width=0.5\textwidth]{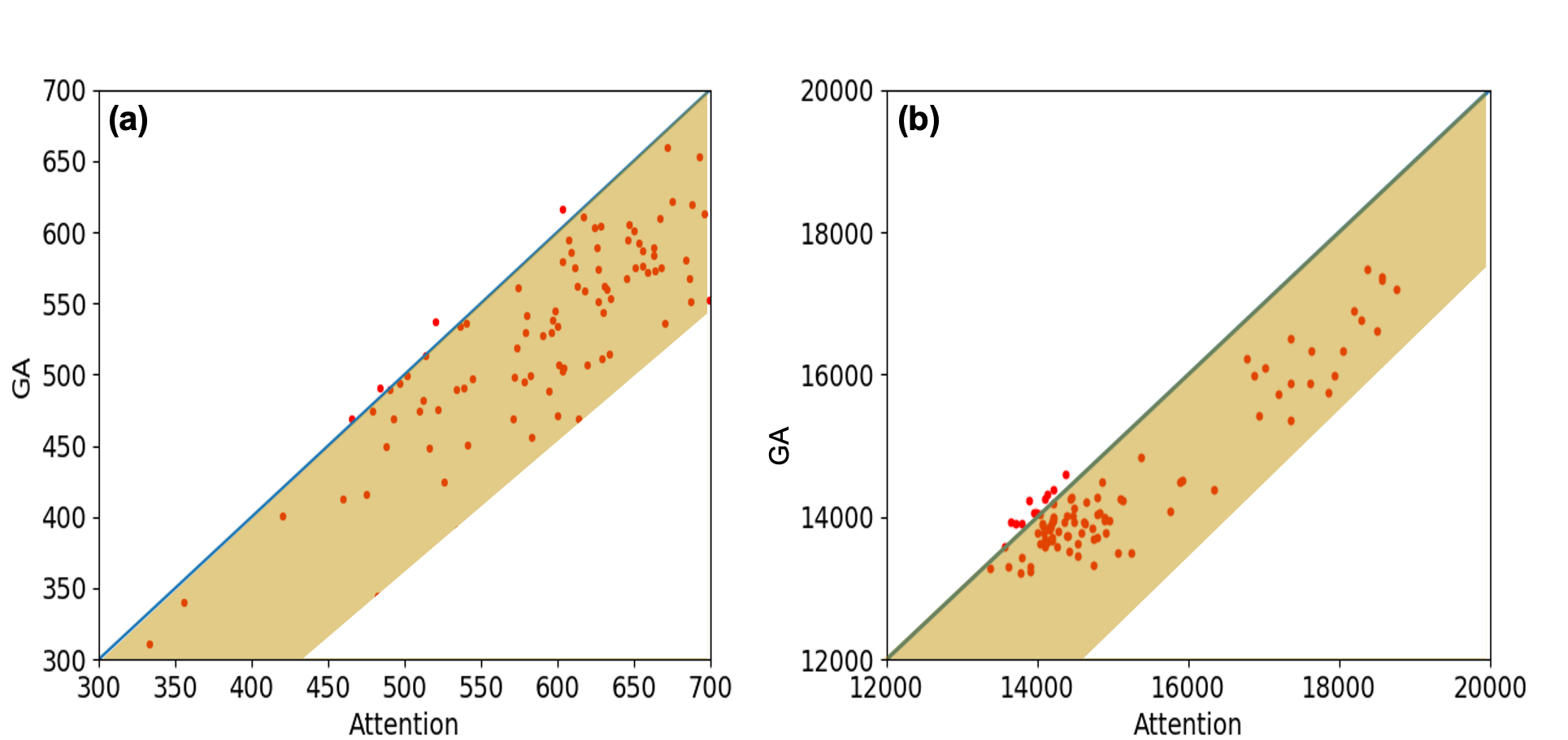}
\caption{Cost comparison between attention model and GA in (a) \textit{Small500}, (b) \textit{Large500}.}
\label{attention_vs_ga} 
\end{figure}

This significant increase in speed enabled by the attention router  is due to different algorithmic structures of the attention model and GA. For the attention model, once the model's training is completed on the training set in an off-line setting, it is applied to new problems in a forward fashion through primarily  matrix multiplications without iterations. The genetic router, on the other hand, solves each problem anew, without the ability to learn from previously solved problems. The significant run-time acceleration enabled by the attention router provides a new alternative for the GA router especially in the early stages of the IC design where placement decisions are yet to be made in the upstream of the workflow. In such instances, the inner optimization involving detailed routing can be significantly accelerated using the attention model as a way to provide useful guidance to the placement algorithm, by leveraging the positive correlations between results from the attention model and GA (Fig.~\ref{attention_vs_ga}). However, as our results suggest, for the ultimate detailed routing decisions, the genetic router currently provides better quality solutions (Fig.~\ref{cost_compare}).  

The routability prediction is crucial in the placement step of IC physical design \cite{zhou2015accurate, chan1993routability}. In order to achieve successful design of a chip, there always exists the need in placement step to fast and accurately assess whether there is a good routing solution exists based on certain placement solution. Existing routability prediction algorithms \cite{zhou2015accurate, chan1993routability,xie2018routenet,brown1993stochastic} have been mainly focusing on global routing stage, and even those that takes into account detailed routing stage \cite{zhou2015accurate}, supervised learning method is used, which makes it depend on other routers to provide labelled data. The attention router in this research provides a promising way for routability prediction by leveraging its positive correlations with genetic router solutions, which is a feasible solution that can be utilized for providing high quality solutions for detailed routing. Another advantage is that, since the attention router utilizes RL, it does not rely on any supervised learning requiring labelled training data. Yet, in order to assess the accuracy of routability prediction with the use of attention router, the model needs to be tested on more problems across different problem sets, which will be part of our future work. 
\subsubsection{Effect of Training Sample Numbers on the Performance of Attention Model.}
\begin{figure}[h]
\includegraphics[width=0.5\textwidth]{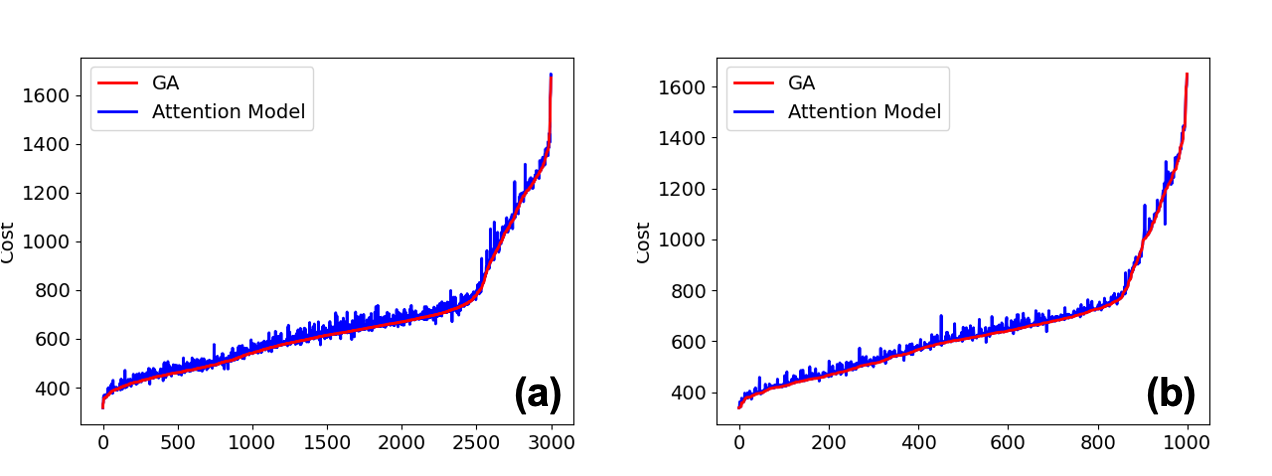}
\caption{Cost comparison with increased number of training problems : (a) \textit{Small5000} training, (b) \textit{Small5000} test.}
\label{LargeData} 
\end{figure}

\textcolor{black}{To investigate the effect of training sample numbers on the performance of the attention model, the model is trained on the  problem set \textit{Small5000}. Fig.~\ref{LargeData} shows the results of the trained model performance on the training and test sets. Compared to the model trained on \textit{Small500} (Fig.~\ref{cost_compare}a and b), the  solutions produced by the attention router improves significantly, approaching the genetic router's solutions.  The sample efficiency of the attention router remains part of our future work. }

\subsubsection{Final Routes}
\begin{figure}[h]
\includegraphics[width=0.5\textwidth]{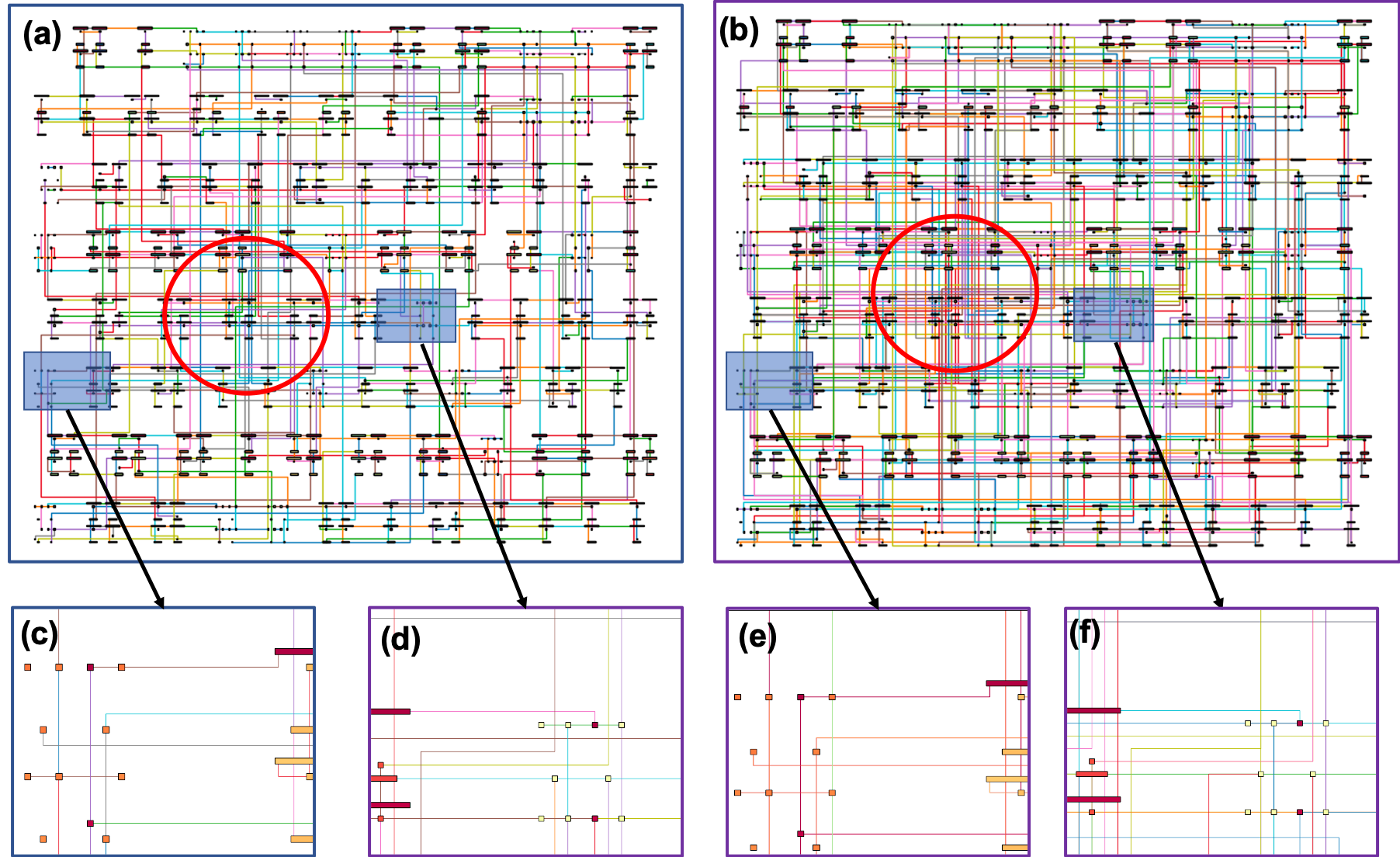}
\caption{Final routes: (a) Attention model and (b) GA on a problem in \textit{Large500} problem set; magnified routes on same regions of (c,d) attention model solution and (e,f) GA solution. }
\label{visualize_route} 
\end{figure}

The routing results  of the attention router and the genetic router on a randomly chosen problem from the \textit{Large500} problem set are shown in Fig.~\ref{visualize_route}. Black dots and bars correspond to instTerms and the colored lines represent the actual routes. As seen, the solutions of the two  routers share some similarity. For instance,  high congestion regions (shown in red circles indicating densely configured routes) are located at similar regions of the physical space. The blow out regions of the attention router (c,d) and the genetic router (e,f) of the same area also indicate similar patterns in routes and similar unconnected instTerms.  This type of similarity suggests  that the attention router can be used by upstream modules to rapidly predict congestion regions as well as  the instTerms that may remain open in the detailed routing stages. 

\section{CONCLUSIONS}

We present a new approach to the track-assignment detailed routing using RL. A detailed routing pipeline we call the attention router that takes into account complex design rule constraints is developed and the attention-model based REINFORCE algorithm is applied for the ordering of instTerm pairs. The attention router and a baseline genetic router is tested on different commercial advanced technologies analog circuits problem sets. The attention router demonstrates a generalization ability to unseen problems from the same problem set after appropriate training. While the genetic router can have slightly better quality  solutions, the attention router is able to achieve more than $100 \times$  acceleration compared to the genetic router without a \textcolor{black}{severe} degradation of the routing solution.  \textcolor{black}{Increasing the number of training problems also greatly improves the performance of the attention router on both training and test sets.} Positive correlations in terms of cost are also found between the attention router and the genetic router, which enable the possibility of applying attention router as a routability predictor in the placement stage. Similarities in the routing solution patterns (congestion region and disconnected instTerms locations) are also discovered, which demonstrate the attention router's ability to work as a more fine-grained congestion predictor and a predictor for disconnected instTerms locations in detailed routing. \textcolor{black}{Future work includes  analyzing the  correlation between the attention router performance and the genetic router in terms of cost and congestion. Sample efficiency of the model will also be studied to provide guidance on training set size when solving problems of different sizes. }

\section{ACKNOWLEDGEMENTS}
This work is funded by the DARPA IDEA program (HR0011-18-3-0010; Funder ID: 10.13039/100006502). The authors would like to thank Prof. Barnabas Poczos for his useful feedback. 



\bibliographystyle{asmems4}
\bibliography{asme2e}

\end{document}